\documentclass{bmvc2k}

\usepackage{amsmath}
\usepackage{amssymb}
\usepackage{subfigure}
\usepackage{caption}
\usepackage{multirow}
\usepackage{array}
\usepackage{algorithm}
\usepackage{algorithmic}
\usepackage[percent]{overpic}
\usepackage{color}
\usepackage{soul}
\newcommand{\R}{\mathbb{R}}

\newcommand{\set}[1]{\mathcal{#1}}

\newcommand{\setI}{\set{I}}

\newcommand{\setK}{\set{K}}
\newcommand{\setL}{\set{L}}

\newcommand{\setN}{\set{N}}

\graphicspath{{./figures/Latex_pictures/}{./figures/GAN_results/}{./figures/generated/}{./figures/PG_GAN_results/}{./figures/sample_blurs/}}

\title{Blind Image Deconvolution using Pretrained Generative Priors}

\addauthor{Muhammad Asim* \\ Fahad Shamshad* \\ Ali Ahmed}{{msee16001, fahad.shamshad, ali.ahmed}@itu.edu.pk}{1}

\addinstitution{
 Department of Electrical Engineering, \\
 Information Technology University, \\
 Lahore, Pakistan
}

\runninghead{Asim, Fahad, Ali}{Blind Image Deconvolution using Generative Priors}

\begin{document}
\maketitle

\begin{abstract}
This paper proposes a novel approach to regularize the
\textit{ill-posed} blind image deconvolution (blind image
deblurring) problem using deep generative networks. We
employ two separate deep generative models --- one trained
to produce sharp images while the other trained to generate
blur kernels from lower-dimensional parameters. To deblur,
we propose an alternating gradient descent scheme operating in the latent lower-dimensional space of each of the
pretrained generative models. Our experiments show excellent deblurring results even under large blurs and heavy
noise. 
To improve the performance on rich image datasets not \textit{well learned} by the generative networks, we present a modification of the proposed scheme that governs the deblurring process under both generative and classical priors.
\end{abstract}

\section{Introduction and Related Work}\label{sec:introduction}
Blind image deblurring aims to recover a true image $i$ and a blur kernel $k$ from blurry and possibly noisy observation $y$. For a uniform and spatially invariant blur, it can be mathematically formulated as 
\begin{equation} \label{eq:bid}
y = i \otimes k + n,
\end{equation}
where $\otimes $ is a convolution operator and $n$ is an additive Gaussian noise. In its full generality, the inverse problem \eqref{eq:bid}  is severely ill-posed as many different instances of $i$, and $k$ fit the observation $y$ \cite{campisi2016blind,kundur1996blind}.

To resolve between multiple instances, priors are introduced on images and/or blur kernels in the image deblurring algorithms. Priors assume an \textit{a priori} model on the true image/blur kernel or both. These natural structures expect images or blur kernels to be sparse in some transform domain; see, for example, \cite{chan1998total, fergus2006removing, levin2009understanding,hu2010single, zhang2011sparse,cai2009blind}. Some of the other penalty functions to improve the conditioning of the blind image deblurring problem are low-rank \cite{ren2016image}, and total variation based priors \cite{pan2014motion}. A recently introduced dark channel prior \cite{pan2016blind} also shows promising results; it assumes a sparse structure on the dark channel of the image, and exploits this structure in an optimization program \cite{xu2011image} to solve the blind image deblurring problem. Other works include extreme channel priors \cite{yan2017image}, outlier robust deblurring \cite{dong2017blind}, learned data fitting \cite{pan2017learning}, and discriminative prior based blind image deblurring approaches \cite{li2018learning}. Although generic and applicable to multiple applications, these engineered models are not very effective as many unrealistic images also fit the prior model \cite{hand2017global}. 

\begin{figure}[t]
\centering
\raisebox{0.1in}{\rotatebox[origin=t]{90}{\footnotesize \hspace{2em} Blurry}} \hspace{-0.0em}
\subfigure{\includegraphics[width=0.17\textwidth]{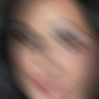}}
\subfigure{\includegraphics[width=0.17\textwidth]{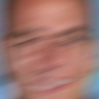}}
\subfigure{\includegraphics[width=0.17\textwidth]{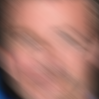}}
\subfigure{\includegraphics[width=0.17\textwidth]{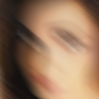}}
\subfigure{\includegraphics[width=0.17\textwidth]{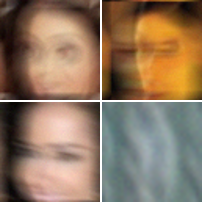}}\\[-1.0em]
\raisebox{0.1in}{\rotatebox[origin=t]{90}{\footnotesize \hspace{2em} Ours}} \hspace{0.15em}
\subfigure{\includegraphics[width=0.17\textwidth]{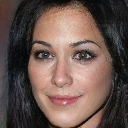}}
\subfigure{\includegraphics[width=0.17\textwidth]{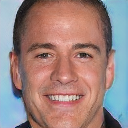}}
\subfigure{\includegraphics[width=0.17\textwidth]{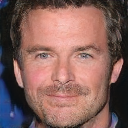}}
\subfigure{\includegraphics[width=0.17\textwidth]{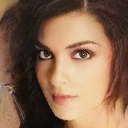}}
\subfigure{\includegraphics[width=0.17\textwidth]{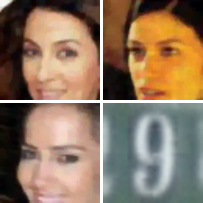}}\\[-0.9em]
\raisebox{0.1in}{\rotatebox[origin=t]{90}{\footnotesize \hspace{2em} Original}} \hspace{-0.0em}
\subfigure{\includegraphics[width=0.17\textwidth]{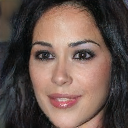}}
\subfigure{\includegraphics[width=0.17\textwidth]{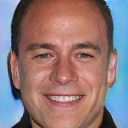}}
\subfigure{\includegraphics[width=0.17\textwidth]{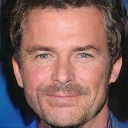}}
\subfigure{\includegraphics[width=0.17\textwidth]{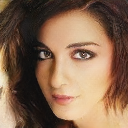}}
\subfigure{\includegraphics[width=0.17\textwidth]{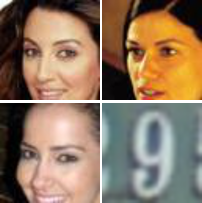}}
\caption{ \small  Blind image deblurring using deep generative priors.}
\label{fig:intro-results}
\end{figure}

Recently deep learning based blind image deblurring approaches have shown impressive results due to their power of learning from large training data \cite{hradivs2015convolutional,nah2016deep,schuler2016learning,xu2017learning,nimisha2017blur,kupyn2017deblurgan}. Generally, these deep learning based approaches invert the forward acquisition model of blind image deblurring via end-to-end training of deep neural networks in a supervised manner. The main drawback of this end-to-end deep learning approach is that it does not explicitly take into account the knowledge of forward map 
\eqref{eq:bid}, but rather learns implicitly from training data. Consequently, the deblurring is more sensitive to changes in the blur kernels, images, or noise distributions in the test set that are not representative of the training data, and often requires expensive retraining of the network for a competitive performance \cite{lucas2018using}. 


Meanwhile, neural network based implicit generative models such as generative adversarial networks (GANs) \cite{goodfellow2014generative} and variational autoencoders (VAEs) \cite{kingma2013auto}  have found much success in modeling complex data distributions especially that of images. Recently, GANs and VAEs have been used for blind image deblurring but only in an end-to end manner \cite{xu2017learning,nimisha2017blur,kupyn2017deblurgan} , which is completely different from our approach as will be discussed in detail. These methods show competitive performance, but since these generative model based approaches are end-to-end they suffer from the same draw backs as other deep learning based debluring approaches. On the other hand, pretrained generative models have recently been employed as regularizers to solve inverse problems in imaging including compressed sensing \cite{bora2017compressed,shah2018solving}, image inpainting \cite{yeh2017semantic}, Fourier ptychography \cite{shamshad2018deep}, and phase retrieval \cite{shamshad2018robust,hand2018phase}. However the applicability of these pretrained generative models in blind image deblurring is relatively unexplored. 

Recently \cite{gandelsman2018double} employ a combination of multiple untrained deep generative models and show their effectiveness on various image layer decomposition tasks including image water mark removal, image dehazing, image segmentation, and transparency separation in images and videos. Different from their approach, we show the effectiveness of our blind image deblurring method by leveraging trained generative models for images and blurs.

In this work, we use the expressive power of pretrained GANs and VAEs to tackle the challenging problem of  blind image deblurring. Our experiments in Figure \ref{fig:intro-results} confirm that integrating deep generative priors in the image deblurring problem enables a far more effective regularization yielding sharper and visually appealing deblurred images. Specifically, our main contributions are
\begin{itemize}
\item To the best of our knowledge, this is the first instance of utilizing pretrained generative models for tackling challenging problem of blind image deblurring.
\item We show that simple gradient descent approach assisted with generative priors is able to recover true image and blur kernel, to with in the range of respective generative models, from blurry image.
\item 
We investigate a modification of the loss function to allow the recovered image some leverage/slack to deviate from the range of the image generator. This modification effectively addresses the performance limitation due to the range of the generator. 

\item Our experiments demonstrate that our approach produce superior results when compared with traditional image priors and unlike deep learning based approaches does not require expensive retraining for different noise levels. 
\end{itemize}

\section{Problem Formulation and Proposed Solution}\label{sec:Problem-Formulation}

We assume the image $i \in \R^n$ and blur kernel $k \in \R^n$ in \eqref{eq:bid} are members of some structured classes $\mathcal{I}$ of images, and $\mathcal{K}$ of blurs, respectively. For example, $\setI$ may be a set of celebrity faces and $\setK$ comprises of  motion blurs. A representative sample set from both classes $\setI$ and $\setK$ is employed to train a generative model for each class. We denote $G_{\mathcal{I}}: \mathbb{R}^l \rightarrow \mathbb{R}^n$ and $G_{\mathcal{K}}: \mathbb{R}^m \rightarrow \mathbb{R}^n$ as the generators for class $\setI$, and $\setK$, respectively. Given low-dimensional inputs $z_i \in \mathbb{R}^l$, and $z_k \in \mathbb{R}^m$, the pretrained generators $G_{\setI}$ and $G_{\setK}$ generate new samples $G_{\setI}(z_i)$, and $G_{\setK}(z_k)$ that are representative of the classes $\setI$ and $\setK$, respectively. Once trained, the weights of the generators are fixed. To recover the sharp image and blur kernel $(i,k)$ from the blurred image $y$ in \eqref{eq:bid}, we propose minimizing the following objective function 
\begin{align}\label{eq:Optimization-Ambient}
(\hat{i},\hat{k}) :=  \underset{\substack{i \in\text{Range}(G_{\setI}) \\k \in\text{Range}(G_{\setK})}}{\text{argmin}} \ \|y - i \otimes k \|^2, 
\end{align}
where $\|\cdot\|$ is the $\ell_2$-distance, Range($G_\setI$) and Range($G_\setK$) is the set of all the images and blurs that can be generated by $G_\setI$ and $G_\setK$, respectively. In words, we want to find an image $i$ and a blur kernel $k$ in the range of their respective generators, that best explain the forward model \eqref{eq:bid}. Ideally, the range of a pretrained generator comprises of only the samples drawn from the probability distribution of the training image or blur class. Constraining the solution $(\hat{i},\hat{k})$ to lie only in generator ranges forces the solution to be the members of classes $\setI$ and $\setK$.

The minimization program in \eqref{eq:Optimization-Ambient} can be equivalently formulated in the lower dimensional, latent representation space as follows:
\begin{align}\label{eq:Optimization-latent}
(\hat{z}_i, \hat{z}_k) = \underset{z_i \in \R^l, z_k \in \R^m}{\text{argmin}}
\ 
\| y - G_{\mathcal{I}}(z_i) \otimes G_\setK(z_k) \|^2.
\end{align}
This optimization program can be thought of as tweaking the latent representation vectors $z_i$ and $z_k$, (input to the generators $G_{\setI}$, and $G_{\setK}$, respectively)  until these generators generate an image $i$ and blur kernel $k$ whose convolution comes as close to $y$ as possible. 
 Incorporating the fact that latent representation vectors $z_i$, and $z_k$ are assumed to be coming from standard Gaussian distributions, 
  we further augment the measurement loss in \eqref{eq:Optimization-latent} with $\ell_2$ penalty terms on the latent representations. The resultant optimization program is then 
\begin{equation} \label{eq:regularized-program}
\underset{z_i \in \R^l, z_k \in \R^m}{\text{argmin}}\ \| y - G_\setI(z_i) \otimes G_\setK(z_k) \|^2+ \gamma\| z_i \|^2 + \lambda\| z_k \|^2,
\end{equation}
where $\gamma$ and $\lambda$ are free scalar parameters. For brevity, we denote the objective function above by $\setL(z_i,z_k)$. Importantly, the weights of the generators are always fixed as they enter into this algorithm as pretrained models. To minimize this non-convex objective, we begin by initializing $z_i$ and $z_k$ by sampling from standard Gaussian distribution, and resort to an alternating gradient descent algorithm by taking a gradient step in one of these while fixing the other to find a minima  $(\hat{z}_i, \hat{z}_k)$. To avoid being stuck in a not good enough local minima, we restart the algorithm with a new random initialization (Random Restarts) when the measurement loss in \eqref{eq:Optimization-latent} does not reduce sufficiently after reasonably many iterations. 
We dubbed proposed deblurring algorithm as \textit{Deep Deblur} and denote blurry image deblurred via \textit{Deep Deblur} as $\hat{\textit{{i}}}_\text{DD}$.
The estimated deblurred image and the blur kernel are acquired by a forward pass of the solutions $\hat{z}_i$ and $\hat{z}_k$ through the generators $G_{\mathcal{I}}$ and $G_{\mathcal{K}}$. Mathematically, $(\hat{i},\hat{k}) = ( G_\mathcal{I}(\hat{z}_i), G_\mathcal{K}(\hat{z}_k))$.

\begin{figure}[t]
\centering
\includegraphics[width=0.9\textwidth]{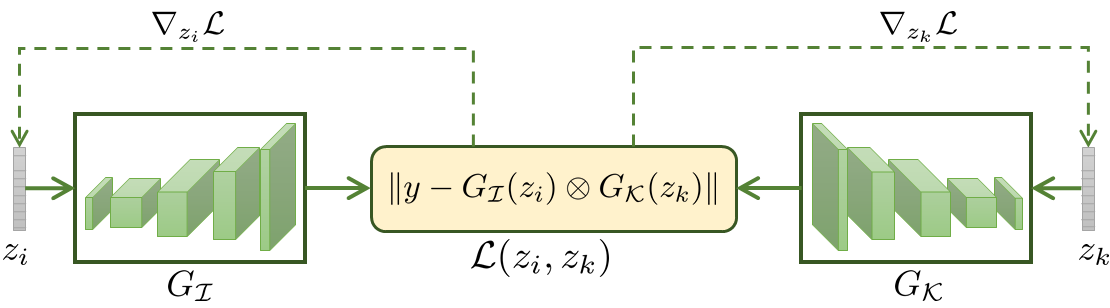}
\caption{\small{ Block diagram of proposed approach. Low dimensional parameters $z_i$ and $z_k$ are updated to minimize the measurement loss using alternating gradient descent. The optimal pair $(\hat{z}_i,\hat{z}_k)$ generate image and blur estimates $(G_\setI(\hat{z}_i), G_\setK(\hat{z}_k))$.}}
\label{fig:proposed_approach}
\end{figure}

\subsection{Beyond the Range of Generator} \label{sec:proposed-method2}
As described earlier, the optimization program \eqref{eq:regularized-program} implicitly constrains the deblurred image to lie in the range of the generator $G_{\setI}$.  This may lead to some artifacts in the deblurred images when the generator range does not completely span the set $\setI$. 
In such case, it makes more sense to not strictly constrain the recovered image to come from the range of the generator, and rather also explore images a bit outside the range. To accomplish this, we propose minimizing the measurement loss of images inside the range exactly as in \eqref{eq:Optimization-latent} together with the measurement loss $\| y - i \otimes G_\setK(z_k) \|^2$ of images not necessarily within the range. The in-range image $G_{\setI}(z_i)$ and the out-range image $i$ are then tied together by minimizing an additional penalty term, $\text{Range Error(i)} := \| i - G_\setI(z_i) \|^2$. The idea is to strictly minimize the range error when pretrained generator has effectively learned the image distribution, and afford some slack otherwise. 
 Finally, to guide the search of a best deblurred image beyond the range of the generator, one of the conventional image priors such as total variation measure $\|\cdot\|_{\text{tv}}$ is also introduced. This leads to the following optimization program 
\begin{align} \label{eq:opt-admm}
\underset{i, z_i, z_k}{\text{argmin}}  \ &
\| y - i \otimes G_\setK(z_k) \|^2 + 
\tau \| i - G_\setI(z_i) \|^2 
 +\zeta\| y - G_\setI(z_i) \otimes G_\setK(z_k) \|^2 + \rho\|i\|_{\text{tv}}
\end{align}
All of the variables are randomly initialized, and the objective is minimized using gradient step in each of the unknowns, while fixing the others. We take the solution $\hat{i}$, and $G(\hat{z}_k)$ as the deblurred image, and the recovered blur kernel. We dubbed this approach as \textit{Deep Deblur with Slack} (DDS) and the image deblurred using this approach is referred to as $\hat{\textit{i}}_\text{DDS}$. 

\section{Experimental Results} \label{experiments}
In this section, we provide a comprehensive set of experiments to evaluate the performance of \textit{Deep Deblur} and \textit{Deep Deblur with Slack} against iterative and deep learning based baseline methods. 
We also evaluate performance under increasing noise and large blurs. In all experiments, we use noisy blurred images, generated by convolving images $i$, and blurs $k$ from their respective test sets and adding 1$\%$ \footnote{For an image scaled between 0 and 1, Gaussian noise of $1\%$ translates to Gaussian noise with standard deviation $\sigma = 0.01$ and mean $\mu=0$.} Gaussian noise (unless stated otherwise). The choice of free parameters in both algorithms for each dataset are provided in the supplementary material.

\subsection{Implementation Details}

\textbf{Datasets}: We choose three image datasets. First dataset, SVHN, consists of house number images from Google street view. A total of 531K images, each of dimension $32\times32 \times 3$, are available out of which 30K are held out as test set. Second dataset, Shoes \cite{yu2014fine} consists of 50K RGB examples of shoes, resized to $64 \times 64 \times 3$. We leave $1000$ images for testing and use the rest as training set. Third dataset, CelebA, consists of relatively more complex images of celebrity faces. A total of 200K, each center cropped to dimension $64 \times 64 \times 3$, are available out of which 22K are held out as a test set. A motion blur dataset is generated consisting of small to very large blurs of lengths varying between 5 and 28; following strategy given in \cite{boracchi2012modeling}. 
We generate 80K blurs out of which 20K is held out as a test set.

\noindent\textbf{Generative Models}: We choose VAE as the generative model for SVHN images and motion blurs. For Shoes and CelebA, the generative model $G_{\setI}$ is the default deep convolutional generative adversarial network (DCGAN) \cite{salimans2016improved}. Further details on architectures of generative models are provided in the supplementary material.

\noindent\textbf{Baseline Methods}: Among the conventional algorithms using engineered priors, we choose dark prior (DP) \cite{pan2016blind}, extreme channel prior (EP) \cite{yan2017image}, outlier handling (OH) \cite{dong2017blind}, and learned data fitting (DF) \cite{pan2017learning} based blind deblurring as baseline algorithms. We optimized the parameters of these methods in each experiment to obtain the best possible baseline results. Among driven approaches for deblurring, we choose \cite{hradivs2015convolutional} that trains a convolutional neural network (CNN) in an end-to-end manner,  and \cite{kupyn2017deblurgan} that trains a neural network (DeblurGAN) in an adversarial manner. 
Deblurred images from these baseline methods will be referred to as $i_\text{DP}$, $i_\text{EP}$, ${i_\text{OH}}$, ${i_\text{DF}}$, $i_\text{CNN}$ and $i_\text{DeGAN}$.

\subsection{Deblurring Results under Pretrained Generative Priors}\label{sec:Exps-PretrainedPriors}

The central limiting factor in the \textit{Deep Deblur} performance is the ability of the generator to represent the (original, clean) image to be recovered. As pointed out earlier that often the generators are not fully expressive (cannot generate new representative samples) on a rich/complex image class such as face images compared to a compact/simple image class such as numbers. Such a generator mostly cannot \textit{adequately} represent a new image in its range. Since \textit{Deep Deblur} strictly constrains the recovered image to lie in the range of image generator, its performance depends on how well the range of the generator spans the image class. Given an arbitrary test image $i_{\text{test}}$ in the set $\setI$, the closest image $i_{\text{range}}$, in the range of the generator, to $i_{\text{test}}$ is computed by solving the following optimization program
\begin{align*}
z_{\text{test}} := \underset{z} {\text{argmin}}  \|i_{\text{test}}-G_{\setI}(z)\|^2,  \quad i_{\text{range}} = G_{\setI}(z_{\text{test}})
\end{align*}
We solve the optimization program by running $10,000(6,000)$ gradient descent steps with a step size of $0.001(0.01)$ for CelebA(SVHN). Parameters for Shoes are the same as CelebA.

A more expressive generator leads to a better deblurring performance as it can well represent an arbitrary original (clean) image $i_{\text{test}}$ leading to a smaller mismatch $ \text{range error} := \|i_{\text{test}} - i_{\text{range}}\|$ to the corresponding range image $i_{\text{range}}$. 

\begin{figure}[t]
\centering
\subfigure[$i$]{\includegraphics[width = 0.15\columnwidth]{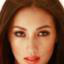}}
\subfigure[$i_\text{range}$]{\includegraphics[width = 0.15\columnwidth]{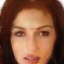}}
\subfigure[$\hat{i}_\text{DD}$]{\includegraphics[width = 0.15\columnwidth]{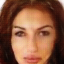}}
\subfigure[$i$]{\includegraphics[width = 0.15\columnwidth]{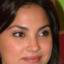}}
\subfigure[$i_\text{range}$]{\includegraphics[width = 0.15\columnwidth]{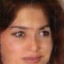}}
\subfigure[$\hat{i}_\text{DD}$]{\includegraphics[width = 0.15\columnwidth]{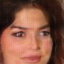}}
\caption{\small  Generator Range Analysis. For each test image $i_\text{test}$ when blurred, \textit{Deep Deblur} tends to recover corresponding range image $i_\text{range}$. 
}
\label{fig:range-images}
\end{figure}

\subsubsection{Impact of Generator Range on Image Deblurring} 
To judge the proposed deblurring algorithms independently of generator range limitations, we present their deblurring performance on range image $i_{\text{range}}$;  we do this by generating a blurred image $y = i_{\text{range}} \otimes k + n$ from an image $i_{\text{range}}$ already in the range of the generator; this implicitly removes the range error as now $i_{\text{test}} = i_{\text{range}}$. We call this range image deblurring, where the deblurred image is obtained using \textit{Deep Deblur}, and is denoted by $\hat{i}_{\text{range}}$. For completeness, we also assess the overall performance of the algorithm by deblurring arbitrary blurred images $y = i_{\text{test}} \otimes k +n$, where $i_{\text{test}}$ is not necessarily in the range of the generator. Unlike above, the overall error in this case accounts for the range error as well. We call this arbitrary image deblurring, and specifically the deblurred image is obtained using \textit{Deep Deblur}, and is denoted by $\hat{i}_\text{DD}$. Figure \ref{fig:range-images} shows a qualitative comparison between $i_{\text{test}}$, $i_{\text{range}}$, and $\hat{i}_\text{DD}$ on CelebA dataset. It is clear that the recovered image $\hat{i}_\text{DD}$ is a good approximation of the range image, $i_{\text{range}}$, 
indicating the limitation of the image generative network. 

\begin{figure}[t]
\centering
\subfigure{\includegraphics[width=0.085\textwidth]{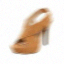}}
\subfigure{\includegraphics[width=0.085\textwidth]{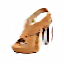}}
\subfigure{\includegraphics[width=0.085\textwidth]{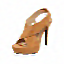}}
\subfigure{\includegraphics[width=0.085\textwidth]{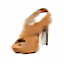}}
\subfigure{\includegraphics[width=0.085\textwidth]{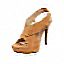}}
\subfigure{\includegraphics[width=0.085\textwidth]{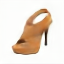}}
\subfigure{\includegraphics[width=0.085\textwidth]{./figures/Shoes/"deblurring - deblurgan1perc  - 1perc noise"/deblurring_41_x_hat.png}}
\subfigure{\begin{overpic}[width=0.085\textwidth]{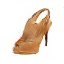}
 \put(72,0){\includegraphics[width = 0.4cm]{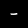}}
\end{overpic}}
\subfigure{\includegraphics[width=0.085\textwidth]{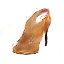}}
\subfigure{\begin{overpic}[width=0.085\textwidth]{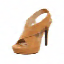}
 \put(72,0){\includegraphics[height = 0.4cm, width = 0.4cm]{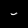}}
\end{overpic}}
\subfigure{\begin{overpic}[width=0.085\textwidth]{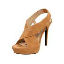}
 \put(72,0){\includegraphics[height = 0.4cm, width = 0.4cm]{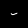}}
\end{overpic}}\\[-0.9em]

\subfigure{\includegraphics[width=0.085\textwidth]{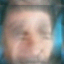}}
\subfigure{\includegraphics[width=0.085\textwidth]{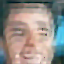}}
\subfigure{\includegraphics[width=0.085\textwidth]{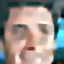}}
\subfigure{\includegraphics[width=0.085\textwidth]{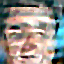}}
\subfigure{\includegraphics[width=0.085\textwidth]{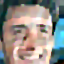}}
\subfigure{\includegraphics[width=0.085\textwidth]{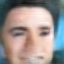}}
\subfigure{\includegraphics[width=0.085\textwidth]{./figures/CelebA/"deblurring - deblurgan1perc  - 1perc noise"/deblurring_19_x_hat.png}}
\begin{overpic}[width=0.085\textwidth]{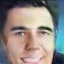}
 \put(64,0){\includegraphics[width = 0.4cm]{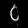}}
\end{overpic}
\subfigure{\includegraphics[width=0.085\textwidth]{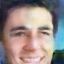}}
\begin{overpic}[width=0.085\textwidth]{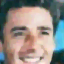}
 \put(64,0){\includegraphics[height = 0.4cm, width = 0.4cm]{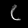}}
\end{overpic}
\begin{overpic}[width=0.085\textwidth]{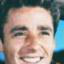}
 \put(64,0){\includegraphics[height = 0.4cm, width = 0.4cm]{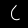}}
\end{overpic}\\[-0.8em]
\setcounter{subfigure}{0}
\subfigure[$y$]{\includegraphics[width=0.085\textwidth]{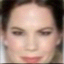}}
\subfigure[$i_{\text{DP}}$]{\includegraphics[width=0.085\textwidth]{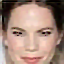}}
\subfigure[$i_{\text{DF}}$]{\includegraphics[width=0.085\textwidth]{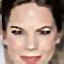}}
\subfigure[$i_{\text{EP}}$]{\includegraphics[width=0.085\textwidth]{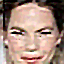}}
\subfigure[$i_{\text{OH}}$]{\includegraphics[width=0.085\textwidth]{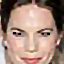}}
\subfigure[$i_{\text{CNN}}$]{\includegraphics[width=0.085\textwidth]{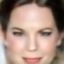}}
\subfigure[\tiny $i_{\text{DeGAN}}$]{\includegraphics[width=0.085\textwidth]{./figures/CelebA/"deblurring - deblurgan1perc  - 1perc noise"/deblurring_38_x_hat.png}}
\subfigure[$\hat{i}_\text{DD}$]{\begin{overpic}[width=0.085\textwidth]{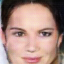}
 \put(64,0){\includegraphics[width = 0.4cm]{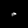}}
\end{overpic}}
\subfigure[${i_\text{range}}$]{\includegraphics[width=0.085\textwidth]{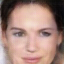}}
\subfigure[$\hat{i}_\text{DDS}$]{\begin{overpic}[width=0.085\textwidth]{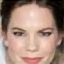}
 \put(64,0){\includegraphics[height = 0.4cm, width = 0.4cm]{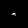}}
\end{overpic}}
\subfigure[$i_{\text{test}}$]{\begin{overpic}[width=0.085\textwidth]{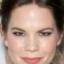}
 \put(64,0){\includegraphics[height = 0.4cm, width = 0.4cm]{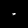}}
\end{overpic}}

\caption{\small{Image deblurring results on Shoes and CelebA using \textit{Deep Deblur} and \textit{Deep Deblur with Slack}. It can be seen that $\hat{i}_\text{DD}$ is in close resemblance to $i_{\text{range}}$ (closest image in the generator range to the original image), where as $\hat{i}_\text{DDS}$ is almost exactly $i_{\text{test}}$, thus mitigating the range issue of image generator.}}
\label{fig:celebA-results}
\end{figure}

\textit{Deep Deblur with Slack} mitigates the range error by not strictly constraining the recovered image to lie in the range of the image generator,
 for details, see  Section \ref{sec:proposed-method2}. As shown in Figure \ref{fig:celebA-results}, estimate $\hat{i}_\text{DDS}$ of true image $i_\text{test}$ from blurry observations is close to $i_\text{test}$ instead of $i_\text{range}$, thus mitigating the range issue.  
 

\begin{figure}
\centering
\subfigure{\includegraphics[width=0.11\textwidth]{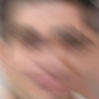}}
\subfigure{\includegraphics[width=0.11\textwidth]{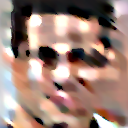}}
\subfigure{\includegraphics[width=0.11\textwidth]{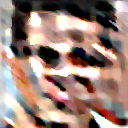}}
\subfigure{\includegraphics[width=0.11\textwidth]{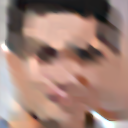}}
\subfigure{\includegraphics[width=0.11\textwidth]{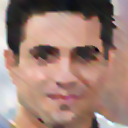}}
\subfigure{\includegraphics[width=0.11\textwidth]{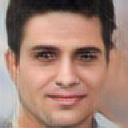}}
\subfigure{\includegraphics[width=0.11\textwidth]{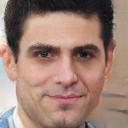}}
\subfigure{\includegraphics[width=0.11\textwidth]{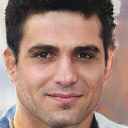}}\\[-0.8em]
\setcounter{subfigure}{0}
\subfigure[$y$]{\includegraphics[width=0.11\textwidth]{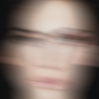}}
\subfigure[$i_\text{DP}$]{\includegraphics[width=0.11\textwidth]{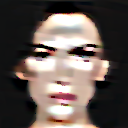}}
\subfigure[$i_\text{EP}$]{\includegraphics[width=0.11\textwidth]{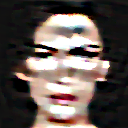}}
\subfigure[$i_\text{DF}$]{\includegraphics[width=0.11\textwidth]{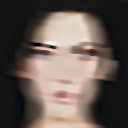}}
\subfigure[$i_\text{OH}$]{\includegraphics[width=0.11\textwidth]{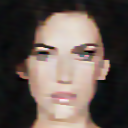}}
\subfigure[$i_\text{DeGAN}$]{\includegraphics[width=0.11\textwidth]{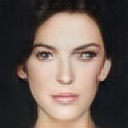}}
\subfigure[$\hat{i}_\text{DD}$]{\includegraphics[width=0.11\textwidth]{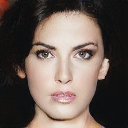}}
\subfigure[$i_\text{sample}$]{\includegraphics[width=0.11\textwidth]{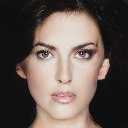}}

\caption{\small{Image deblurring results on blurry images generated from samples, $i_\text{sample}$, of PGGAN via \textit{Deep Deblur}. Visually appealing images, $\hat{i}_\text{DD}$, are recovered,  from blurry ones.}}
\label{fig:pg-gan-results}
\end{figure}

\begin{figure*}[h]
\centering
\subfigure{\includegraphics[width=0.09\textwidth]{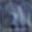}}
\subfigure{\includegraphics[width=0.09\textwidth]{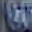}}
\subfigure{\includegraphics[width=0.09\textwidth]{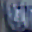}}
\subfigure{\includegraphics[width=0.09\textwidth]{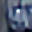}}
\subfigure{\includegraphics[width=0.09\textwidth]{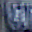}}
\subfigure{\includegraphics[width=0.09\textwidth]{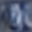}}
\subfigure{\includegraphics[width=0.09\textwidth]{./figures/SVHN/"deblurring - deblurgan1perc  - 1perc noise"/deblurring_92_x_hat.png}}
\begin{overpic}[width=0.09\textwidth]{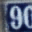}
 \put(55,0){\includegraphics[height = 0.5cm, width = 0.5cm]{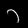}}
\end{overpic}
\subfigure{\includegraphics[width=0.09\textwidth]{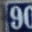}}
\begin{overpic}[width=0.09\textwidth]{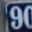}
 \put(55,0){\includegraphics[height = 0.5cm, width = 0.5cm]{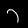}}
\end{overpic}\\[-1.0em]
\setcounter{subfigure}{0}
\subfigure[$y$]{\includegraphics[width=0.09\textwidth]{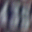}}
\subfigure[$i_{\text{DP}}$]{\includegraphics[width=0.09\textwidth]{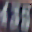}}
\subfigure[$i_{\text{OH}}$]{\includegraphics[width=0.09\textwidth]{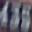}}
\subfigure[$i_{\text{DF}}$]{\includegraphics[width=0.09\textwidth]{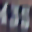}}
\subfigure[$i_{\text{EP}}$]{\includegraphics[width=0.09\textwidth]{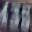}}
\subfigure[$i_{\text{CNN}}$]{\includegraphics[width=0.09\textwidth]{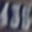}}
\subfigure[$i_{\text{DGAN}}$]{\includegraphics[width=0.09\textwidth]{./figures/SVHN/"deblurring - deblurgan1perc  - 1perc noise"/deblurring_96_x_hat.png}}
\subfigure[$\hat{i}_\text{DD}$]{\begin{overpic}[width=0.09\textwidth]{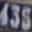}
 \put(55,0){\includegraphics[height = 0.5cm, width = 0.5cm]{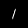}}
\end{overpic}}
\subfigure[${i_\text{range}}$]{\includegraphics[width=0.09\textwidth]{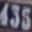}}
\subfigure[$i_{\text{test}}$]{\begin{overpic}[width=0.09\textwidth]{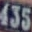}
 \put(55,0){\includegraphics[height = 0.5cm, width = 0.5cm]{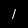}}
\end{overpic}} 

\caption{\small{Image deblurring results on SVHN images using \textit{Deep Deblur}. It can be seen that due to the simplicity of these images, $\hat{i}_\text{DD}$ is a visually a very good estimate of $i_{\text{test}}$, due to the close proximity between $i_\text{range}$ and $i_{\text{test}}$.}}
\label{fig:svhn-results}
\end{figure*}


\begin{figure}[t]
\begin{minipage}{0.45\textwidth}
\centering
\subfigure{\includegraphics[width=0.235\columnwidth]{./figures/SVHN/"deblurring -  6perc noise - 10RR"/deblurring_34_y_from_test.png}}
\subfigure{\includegraphics[width=0.235\columnwidth]{./figures/SVHN/"deblurring -  deblurgan1to10perc - 6perc noise"/deblurring_34_x_hat.png}}
\subfigure{\includegraphics[width=0.235\columnwidth]{./figures/SVHN/"deblurring -  6perc noise - 10RR"/deblurring_34_x_hat_from_test.png}}
\subfigure{\includegraphics[width=0.235\columnwidth]{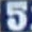}} 
\\[-0.8em]
\setcounter{subfigure}{0}
\subfigure[ $y$]{\includegraphics[width=0.235\columnwidth]{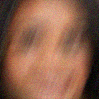}}
\subfigure[ $i_\text{DeGAN}^*$]{\includegraphics[width=0.235\columnwidth]{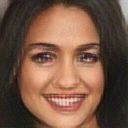}}
\subfigure[ $\hat{i}_\text{DD}$]{\includegraphics[width=0.235\columnwidth]{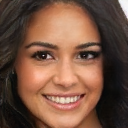}}
\subfigure[ $i_{\text{test}}$]{\includegraphics[width=0.235\columnwidth]{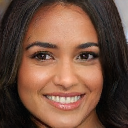}}
\caption{\small Visual Comparison of DeblurGAN ($i_\text{DeGAN}^*$) trained on $1$-$10\%$ noise with \textit{Deep Deblur} on noisy images from SVHN (top row) and samples from PGGAN (bottom row). }
\label{fig:results-heavynoise}
\end{minipage}
\hfill
\begin{minipage}{0.52\textwidth}
\centering
\raisebox{-0.05in}{\rotatebox[origin=t]{90}{\footnotesize \hspace{2.5em} $y$}} \hspace{-0.0em}
\subfigure{\includegraphics[width=0.18\columnwidth]{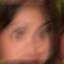}}
\subfigure{\includegraphics[width=0.18\columnwidth]{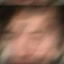}}
\subfigure{\includegraphics[width=0.18\columnwidth]{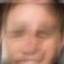}}
\subfigure{\includegraphics[width=0.18\columnwidth]{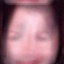}}
\subfigure{\includegraphics[width=0.18\columnwidth]{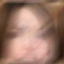}}\\[-1.6em]
\raisebox{-0.05in}{\rotatebox[origin=t]{90}{\footnotesize \hspace{4em} $\hat{i}_\text{DDS}$}} \hspace{-0.35em}
\subfigure{\includegraphics[width=0.18\columnwidth]{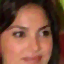}}
\subfigure{\includegraphics[width=0.18\columnwidth]{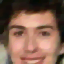}}
\subfigure{\includegraphics[width=0.18\columnwidth]{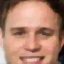}}
\subfigure{\includegraphics[width=0.18\columnwidth]{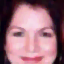}}
\subfigure{\includegraphics[width=0.18\columnwidth]{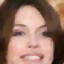}}\\[-2.5em]
\raisebox{-0.05in}{\rotatebox[origin=t]{90}{\footnotesize \hspace{4em} $i_{\text{test}}$}} \hspace{-0.1em}
\subfigure{\includegraphics[width=0.18\columnwidth]{./figures/CelebA/"Original Images"/x_orig_32.png}}
\subfigure{\includegraphics[width=0.18\columnwidth]{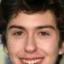}}
\subfigure{\includegraphics[width=0.18\columnwidth]{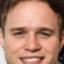}}
\subfigure{\includegraphics[width=0.18\columnwidth]{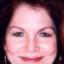}}
\subfigure{\includegraphics[width=0.178\columnwidth]{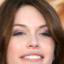}}
\caption{\small Large Blurs. Under large blurs, proposed \textit{Deep Deblur with Slack}, shows excellent deblurring results. }
\label{fig:results-largeblur}
\end{minipage}
\end{figure}

\subsubsection{Deblurring Results on CelebA, Shoes and SVHN}
\textbf{Qualitative results on CelebA}: Figure \ref{fig:celebA-results} gives a qualitative comparison between $i$, $i_{\text{range}}$, $\hat{i}_\text{DD}$, $\hat{i}_\text{DDS}$, and baseline approaches on CelebA and Shoes dataset. 
The deblurred images under engineered priors are qualitatively a lot inferior than the deblurred images $\hat{i}_\text{DD}$, and $\hat{i}_\text{DDS}$ under the proposed generative priors, especially under large blurs. On the other hand, the end-to-end training based approaches CNN, and DeblurGAN perform relatively better, however, the well reputed CNN is still displaying over smoothed images with missing edge details, etc  compared to our results $\hat{i}_\text{DDS}$. DeblurGAN, though competitive, is outperformed by the proposed \textit{Deep Deblur with Slack} by more than 1.5dB as shown in Table \ref{table:psnr-ssim-results}. 
The images $\hat{i}_\text{DD}$ are sharp and with well defined facial boundaries and markers owing to the fact they strictly come from the range of the generator, however, in doing so these images might end up changing some image features such as expressions, nose, etc. On a close inspection, it becomes clear that how well $\hat{i}_\text{DD}$ approximates $i_{\text{test}}$ roughly depends (see, images specifically in Figure \ref{fig:range-images}) on how close $i_{\text{range}}$ is to $i_{\text{test}}$ exactly.   While as $\hat{i}_\text{DDS}$ are allowed some leverage, and are not strictly confined to the range of the generator, they tend to agree more closely with the ground truth. We go on further by utilizing pretrained PGGAN \cite{karras2017progressive} in \textit{Deep Deblur} by convolving sampled images with large blurs ($30 \times 30$); see Figure \ref{fig:pg-gan-results}. It has been observed that pre-trained generators struggle at higher resolutions \cite{athar2018latent}, so we restrict our results at $128 \times 128$ resolution. 
In Figure \ref{fig:pg-gan-results}, it can be seen that under expressive generative priors our approach exceeds all other baseline methods recovering fine details from extremely blurry images.

 \textbf{Qualitative Results on SVHN}: Figure \ref{fig:svhn-results} gives qualitative comparison between proposed and baseline methods on SVHN dataset. Here the deblurring under classical priors again clearly under performs compared to the proposed image deblurring results $\hat{i}_\text{DD}$. CNN also continues to be inferior, and the DeblurGAN also shows artifacts. We do not include the results $\hat{i}_\text{DDS}$ in these comparison as $\hat{i}_\text{DD}$ already comprehensively outperform the other techniques on this dataset. The convincing results $\hat{i}_\text{DD}$ are a manifestation of the fact that unlike the relatively complex CelebA and Shoes datasets, the simpler image dataset SVHN is effectively spanned by the range of the image generator. 

\begin{table}[t]
\centering
\scalebox{0.85}{
\renewcommand{\arraystretch}{1.2}
\begin{tabular}{|c|c|c|c|c|c|c|c|c|c|c|}
\hline
\multicolumn{2}{|c|}{\textbf{Method}} & $\hat{i}_\text{EP}$ & $\hat{i}_\text{DF}$ & $\hat{i}_\text{OH}$ & $\hat{i}_\text{DP}$ & $\hat{i}_\text{DeGAN}$ & $\hat{i}_\text{CNN}$ & $\hat{i}_\text{DD}$ & $\hat{i}_\text{DDS}$ & $\hat{i}_\text{range}$ \\ \hline
\multirow{2}{*}{\textbf{SVHN}} & PSNR & 20.35 & 20.64 & 20.82 & 20.91 & 15.79 & 21.24 & 24.47 & - & 30.13 \\ \cline{2-11} 
 & SSIM & 0.55 & 0.60 & 0.58 & 0.58 & 0.54 & 0.63 & 0.80 & - & 0.89 \\ \hline
\multirow{2}{*}{\textbf{Shoes}} & PSNR & 18.33 & 17.79 & 19.04 & 18.45 & 21.84 & 24.76 & 21.20 & 26.98 & 23.93 \\ \cline{2-11} 
 & SSIM & 0.73 & 0.73 & 0.76 & 0.74 & 0.85 & 0.89 & 0.83 & 0.93 & 0.87 \\ \hline
\multirow{2}{*}{\textbf{CelebA}} & PSNR & 17.80 & 20.00 & 20.71 & 21.09 & 24.01 & 23.75 & 21.11 & 26.60 & 25.49 \\ \cline{2-11} 
 & SSIM & 0.70 & 0.79 & 0.81 & 0.79 & 0.88 & 0.87 & 0.80 & 0.93 & 0.91 \\ \hline
\end{tabular}
\renewcommand{\arraystretch}{1.0}
}\\[0.4em]

\caption{ \small{Quantitative comparison of proposed approach with baseline methods on CelebA, SVHN, and Shoes dataset. Table shows average PSNR and SSIM on 80 random images from respective test sets.}}
\label{table:psnr-ssim-results}
\end{table}

\textbf{Quantitative Results}: Quantitative results for CelebA, Shoes\footnote{For qualitative results, see supplementary material.} and SVHN using peak-signal-to-noise ratio (PSNR) and structural-similarity index (SSIM) \cite{wang2004image}, averaged over 80 respective test set images, are given in Table \ref{table:psnr-ssim-results}. On CelebA and Shoes, the results clearly show a better performance of \textit{Deep Deblur with Slack}, on average, compared to all baseline methods. On SVHN, the results show that \textit{Deep Deblur} outperforms all competitors. The fact that \textit{Deep Deblur} performs more convincingly on SVHN is explained by observing that the range images $i_{\text{range}}$ in SVHN are quantitatively much better compared to range images of CelebA and Shoes. 

\subsubsection{Robustness against Noise and Large Blurs}
\begin{figure}[t]
\centering
\subfigure{\includegraphics[width=1\textwidth]{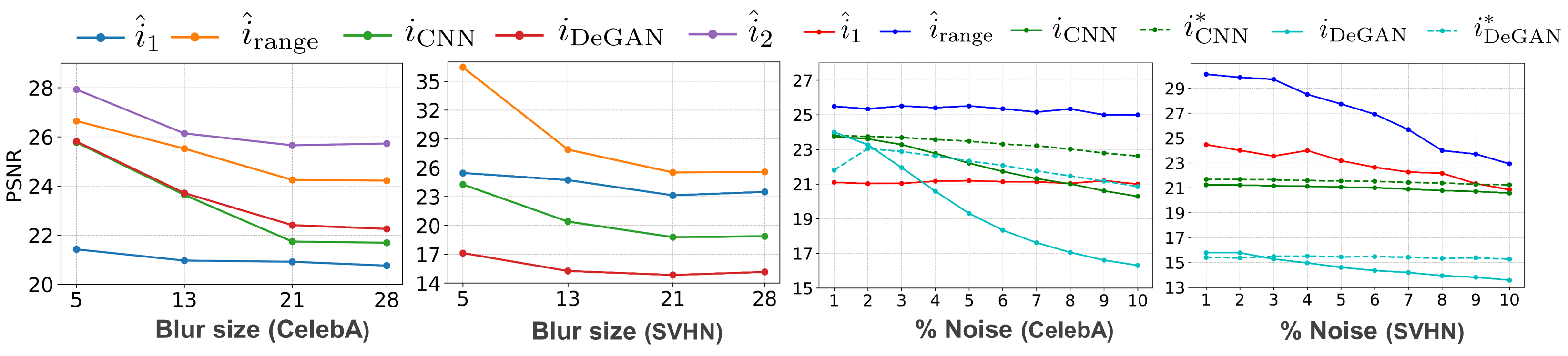}}
\caption{\small Blur Size and Noise Analysis. Comparative performance of proposed methods, on CelebA and SVHN dataset, against baseline techniques, as blur length and noise level increases.}
\label{fig:psnr-ssim-blursize}
\end{figure}

\textbf{Robustness against Noise}: Figure \ref{fig:psnr-ssim-blursize} gives a quantitative comparison of the deblurring obtained via \textit{Deep Deblur} (the free parameters $\lambda$, $\gamma$ and random restarts in the algorithm are fixed as before), and baseline methods CNN, DeblurGAN (trained on fixed 1\% noise level and on varying  1-10\% noise levels) in the presence of Gaussian noise. We also include the performance of deblurred range images $\hat{i}_{\text{range}}$, introduced in Section \ref{sec:Exps-PretrainedPriors}, as a benchmark. Conventional prior based approaches are not included as their performance substantially suffers on noise compared to other approaches. On the vertical axis, we plot the PSNR and on the horizontal axis, we vary the noise strength from 1 to 10\%. In general, the quality of deblurred range images (expressible by the generators) $\hat{i}_{\text{range}}$ under generative priors surpasses other algorithms on both CelebA, and SVHN. This in a way manifests that under expressive generative priors, the performance of our approach is far superior. The quality of deblurred images $\hat{i}_\text{DD}$ under generative priors with arbitrary (not necessarily in the range of the generator) input images is the second best on SVHN, however, it under performs on the CelebA dataset; the most convincing explanation of this performance deficit is the range error (not as expressive generator) on the relatively complex/rich images of CelebA. The end-to-end approaches trained on fixed 1\% noise level display a rapid deterioration on other noise levels. Comparatively, the ones trained on 1-10\% noise level, expectedly, show a more graceful performance. Qualitative results under heavy noise are depicted in Figure \ref{fig:results-heavynoise}. Our deblurred image $\hat{i}_\text{DD}$ visually agrees better with $i_{\text{test}}$ than other methods. 

\textbf{Robustness against Large Blurs}: 
Figure \ref{fig:results-largeblur} shows the deblurred images obtained from a very blurry face image. The deblurred image $\hat{i}_\text{DDS}$ using \textit{Deep Deblur with Slack} is able to recover the true face from a completely unrecognizable face. The classical baseline algorithms totally succumb to such large blurs. The quantitative comparison against end-to-end neural network based methods CNN, and DeblurGAN is given in Figure \ref{fig:psnr-ssim-blursize}. We plot the blur size against the average PSNR for both Shoes, and CelebA datasets. On both datasets, deblurred images $\hat{i}_\text{DDS}$ convincingly outperforms all other techniques. For comparison, we also add the performance of  $\hat{i}_{\text{range}}$. Excellent deblurring under large blurs can also be seen in Figure \ref{fig:pg-gan-results} for PGGAN. To summarize, the end-to-end approaches begin to lag a lot behind our proposed algorithms when the blur size increases. This is owing to the firm control induced by the powerful generative priors on the deblurring process in our newly proposed algorithms.

\section{Conclusion} \label{sec:conc}

This paper proposes a novel framework for blind image deblurring that uses deep generative networks as priors rather than in a conventional end-to-end manner. We report convincing deblurring results under the generative priors in comparison to the existing methods. A thorough discussion on the possible limitations of this approach on more complex images is presented along with an effective remedy to address these shortcomings. Our main contribution lies in introducing pretrained generative model in solving blind deconvolution. Introducing more expressive generative models with new novel architectures would improve the performance. We leave these exciting directions for future work. 


\bibliographystyle{ieeetr}
\bibliography{egbib}
\end{document}